\documentclass{article}

\usepackage[final]{paperstyle2024}

\usepackage[utf8]{inputenc} 
\usepackage[T1]{fontenc}    
\usepackage{hyperref}       
\usepackage{url}            
\usepackage{booktabs}       
\usepackage{amsfonts}       
\usepackage{nicefrac}       
\usepackage{microtype}      
\usepackage{xcolor}         

\usepackage{subcaption}      

\usepackage{booktabs}
\usepackage{tikz}
\usepackage{pifont} 
\usepackage{amsfonts}
\usepackage{amsmath}

\newcommand{\cut}[1]{}

\usepackage{enumitem}
\usepackage{comment}

\usepackage{algpseudocode}
\usepackage{algorithm}

\title{Combining LLM decision and RL action selection\\to improve RL policy for adaptive interventions}

\author{%
  Karine Karine \\
  University of Massachusetts Amherst, USA \\
  \texttt{karine@cs.umass.edu} \\
  \And
  Benjamin M. Marlin \\
  University of Massachusetts Amherst, USA \\
  \texttt{marlin@cs.umass.edu} \\
}

\begin{document}

\maketitle

\begin{abstract}
Reinforcement learning (RL) is increasingly being used in the healthcare domain, particularly for the development of personalized health adaptive interventions. Inspired by the success of Large Language Models (LLMs), we are interested in using LLMs to update the RL policy in real time, with the goal of accelerating personalization. We use the text-based user preference to influence the action selection on the fly, in order to immediately incorporate the user preference. We use the term ``user preference'' as a broad term to refer to a user personal preference, constraint, health status, or a statement expressing like or dislike, etc. Our novel approach is a hybrid method that combines the LLM response and the RL action selection to improve the RL policy. Given an LLM prompt that incorporates the user preference, the LLM acts as a filter in the typical RL action selection. We investigate different prompting strategies and action selection strategies. To evaluate our approach, we implement a simulation environment that generates the text-based user preferences and models the constraints that impact behavioral dynamics. We show that our approach is able to take into account the text-based user preferences, while improving the RL policy, thus improving personalization in adaptive intervention.
\end{abstract}

\section{Introduction}
\label{sec: LLM RL Introduction}

Reinforcement learning (RL) is increasingly being used in the healthcare domain, particularly for the development of personalized health adaptive interventions \citep{coronato2020reinforcement, yu2021reinforcement, gonul2021reinforcement, liao2020personalized}. 
%
%
Inspired by the success of Large Language Models (LLMs), we are interested in using LLMs to update the RL policy in real time, with the goal of accelerating personalization.
We use the text-based user preference to influence the selection of actions on the fly, in order to immediately incorporate the user preference. We use the term ``user preference'' as a broad term to refer to a user personal preference, constraint, health status, or a statement expressing like or dislike, etc. 

We illustrate our motivation with an example from the behavioral domain, where researchers study the effectiveness of a mobile health app that encourages positive behavior change (i.e., exercise more or reduce smoking) by sending messages to the user, a.k.a participant \citep{nahum2018just, hardeman2019systematic}. Often, there can be too many messages sent to the user, or some issues in the decision rule or policy that result in incorrectly contextualized messages sent to the user (e.g., when the user preference does not align with the policy). These messages may annoy the user, or even cause the user to disengage from the study. Thus, it is critical to take into account the user preference before it becomes too late or irreversible (e.g. the user exits the study). 

One solution is to allow the user to specify their preferences in the form of free-text descriptions, and immediately take them into account to influence the action selection. This is especially relevant in today's generation, where people use chats and social media to communicate.
For example, the user preference can be: ``I sprained my ankle'', ``I don't want messages during the weekend'' or ``I don't like to receive negative messages about smoking''.

Our goal is to immediately incorporate the user preferences, by using LLMs, in order to improve the RL policy,
and thus accelerate personalization in adaptive intervention.

However, implementing methods using LLMs comes with challenges: (1) how to construct effective LLM prompts to obtain the desired LLM response, (2) how to incorporate the LLM response in the RL system, in order to update the RL policy, and (3) how to evaluate the new method, since incorporating the user preference introduces additional constraints on the behavioral dynamics.

Thus, our problem statement is: ``How can we use LLMs to update the RL policy to accelerate personalization in adaptive interventions?''. We answer this question by introducing: (1) our novel hybrid method ``LLM+TS'', (2) a novel simulation environment ``StepCountJITAI for LLM'' that works with LLMs and allows for the evaluation of our new method, and (3) a solution framework for implementing a pipeline for personalized health adaptive
interventions.

\vspace{1em}

\noindent \textbf{Our contributions are}:

\begin{enumerate}

    \item \textbf{LLM+TS: combining LLM decision and RL action selection to improve RL policy.} We introduce our novel hybrid method called ``LLM+TS'', that combines the LLM response and the RL action selection to improve the RL policy for adaptive intervention. We use Thompson Sampling (TS) as the RL agent, because it is an effective Bayesian approach that requires fewer iterations than typical deep RL methods \citep{agrawal2013}. 

    Our approach incorporates prompting strategies and action selection strategies to accelerate personalization. For our approach, we implement two loops: (1) a simulation environment that generates text-based user preferences, which we describe below, and (2) the typical RL loop where the RL agent selects a candidate action at each time step. Then, based on the LLM prompt that includes the user preference and other information, the LLM decides whether to ``not send'' or ``send'' a message. The LLM acts as a filter in the typical RL action selection.
    
    \item \textbf{StepCountJITAI for LLM: novel simulation environment that generates user preferences and incorporates constraints to impact behavioral dynamics.} We extend the base simulator for adaptive intervention, introduced in \cite{karine2024}, to create a new simulation environment that works with LLMs.    
    StepCountJITAI for LLM is used to evaluate our new method. To construct it, we: (1) augment the user state with an auxiliary variable with dynamics that follow a Markov chain, (2) generate the text-based user preference based on the auxiliary variable value, (3) incorporate constraints that impact the behavioral dynamics implemented inside the simulation environment (e.g., if the user is in a ``cannot walk'' state, then the reward value drops to $0$, the disengagement risk increases, and the habitual level increases). We note that our new process for generating the text-based user preference is separate from the RL loop, and is only included in the simulation environment. Our StepCountJITAI for LLM is an innovative simulation environment that has not been explored in prior work, and it offers significant potential for advancing the development of new RL algorithms for adaptive interventions using LLM. StepCountJITAI is available at: github.com/reml-lab/StepCountJITAI.
    
    \item \textbf{Practical method for accelerating personalization in adaptive intervention.} We demonstrate how to frame a physical activity adaptive intervention as an RL system using LLMs. We show that our approach improves the RL policy while incorporating the user preference. Our method offers a promising solution framework for implementing a pipeline for personalized health adaptive intervention. 
\end{enumerate}

\begin{figure}[th]
\centering
\includegraphics[width=0.55\columnwidth]{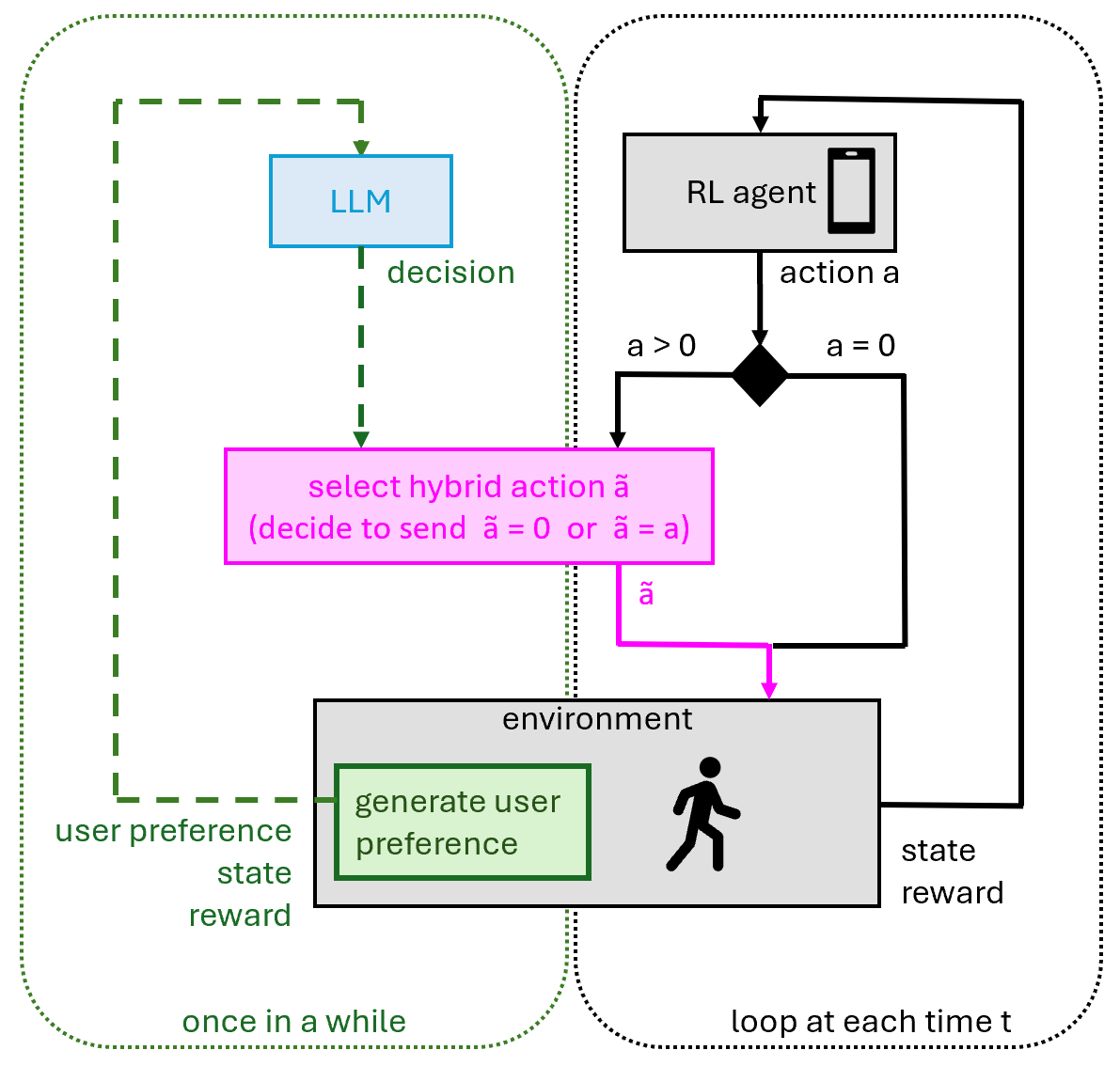} 
\caption{Overview of LLM+TS method. LLM+TS is a hybrid method that combines LLM decision and RL action
selection to improve the RL policy. The LLM prompt includes information such as a description of the behavioral dynamics, current user state and some past data, user preference (constraint) and a question asking the LLM to decide: ``not send'' or ``send'' a message (i.e., $\tilde{a}=0$ or $\tilde{a}=a$). The LLM acts as a filter in the typical RL action selection.
}
\label{fig: Overview}
\end{figure}

\noindent We provide an overview of our novel method in Figure \ref{fig: Overview}, and details in Section \ref{sec: LLM RL Methods}.

\newpage

\vspace{1em}

\section{Background}
\label{sec: LLM RL Background}

\label{sec: background StepCountJITAI}
\noindent\textbf{StepCountJITAI: simulation environment for adaptive intervention.} There is limited prior work on simulation environments for adaptive intervention in the literature. In this work, we extend the base simulator for adaptive intervention introduced in \cite{karine2024}. This base simulator was specifically designed to be used for the development of new RL algorithms for adaptive intervention. 

A physical activity adaptive intervention can be framed as an RL system, where the types of messages are the possible actions. In our simulation environment, we use the following values: $a=0$ (do not send a message), $a=1$ (send a generic message), $a=2$ (send a message tailored to context $0$), and $a=3$ (send a message tailored to context $1$). 

The context can be, for example, a binary state of the participant: `stressed / not stressed' or `at home / at work' or `smoker / not a smoker', etc. Note that the context can be extended to include multiple values.

The environment states include the participant behaviors: habituation level and disengagement risk.

For the notation, we use an uppercase letter for the variable name, and a lowercase letter for the variable value, for example: the context variable $C$ has value $c_t =0$ at time $t$.

Below we describe some of the simulation environment variables and parameters that are used in the behavioral dynamics:
$c_t$ is the true context, $p_{t}$ is the probability of context $1$, $l_t$ is the inferred context, $h_t$ is the habituation level, $d_t$ is the disengagement risk, $s_t$ is the step count ($s_t$ is the participant's number of walking steps), and $a_{t}$ is the action at time $t$. The base simulator also includes behavioral parameters: $\delta_d$ and $\epsilon_d$ are decay and increment parameters for the disengagement risk, and $\delta_h$ and $\epsilon_h$ are decay and increment parameters for the habituation level.

The goal is to increase the participant's walking step count. Thus, the walking step count is also the RL reward. 

This base simulator implements complex behavioral dynamics. The behavioral dynamics can be summarized as follows: Sending a message causes the habituation level to increase. Not sending a message causes the habituation level to decrease. An incorrectly tailored message causes the disengagement risk to increase. A correctly tailored message causes the disengagement risk to decrease. When the disengagement risk exceeds a given threshold, the behavioral study ends. The reward is the surplus step count, beyond a baseline count, attenuated by the habituation level. These behavioral dynamics can be translated into equations, which we present below. 
%
%
\label{Section Background: behavioral dynamics}
\begin{small}
\begin{align}
        c_{t+1} &\sim \mathit{Bernoulli}(0.5), \;\;\; x_{t+1} \sim \mathcal{N}(c_{t+1}, \sigma^2)\\
        p_{t+1} &= P(C=1|x_{t+1}), \;\;\; l_{t+1} = p_{t+1} >0.5\\
        %
    %
        h_{t+1} &=   \begin{cases}
                    (1-\delta_h) \cdot  h_{t}             &\text{~~if~} a_{t} = 0\\
                    \text{min}(1, h_{t} + \epsilon_h)     & \text{~~otherwise}\\
                \end{cases}\\
        d_{t+1} &=   \begin{cases}
                    d_{t}                                 &\text{~~if~} a_{t} = 0\\
                    (1-\delta_d) \cdot  d_{t}             &\text{~~if~} a_{t} \in \{1,c_{t}+2\}\\
                    \text{min}(1, d_{t} + \epsilon_d)     &\text{~~otherwise}
                \end{cases}
\end{align}
\end{small}
\begin{small}
\begin{align}
    s_{t+1} &=   \begin{cases}
                m_{s}    + (1-h_{t+1}) \cdot  \rho_1  &\text{~~~~~~~~~~if~} a_{t} = 1\\
                m_{s}    + (1-h_{t+1}) \cdot  \rho_2  &\text{~~~~~~~~~~if~} a_{t} = c_{t}+2\\
                m_{s}    & \text{~~~~~~~~~~otherwise}
            \end{cases}
\end{align}
\end{small}


where $\sigma$ is the context uncertainty, $x_t$ is the context feature, $\sigma, \rho_1, \rho_2, m_s$ are fixed parameters. We use the same default parameter values as the base simulator, which we summarize in Appendix \ref{Appendix: JITAI simulation environment specifications}.

However this base simulator does not support LLMs. Thus, we extend this base simulator to create our novel simulation environment that includes the support for LLMs and the constraints arising from inserting the user preference. We describe our novel simulation environment in Section \ref{sec: create StepCountJITAI for LLM}.

\vspace{1em}

\label{Section Background: Thompson Sampling}
\noindent\textbf{Thompson Sampling.} 
Thompson Sampling (TS) is a probabilistic method for decision-making under uncertainty. It can be used to address contextual multi-armed bandit problems \citep{russo2018,  agrawal2013, chu2011contextual, Thompson1933}. Typical TS for contextual bandit settings uses a reward model of the form $\mathcal{N}(r;\theta_a^\top \mathit{v}_t,\sigma_{Ya}^2)$, where $\mathit{v}_t$ is the state vector at time $t$, $\theta_a$ is a vector of weights, and $\sigma_{Ya}^2$ is the reward variance for action $a$. Thus, $\theta_a^\top \mathit{v}_t$ represents the mean reward for action $a$.

The reward model weights $\theta_a$ are random variables of the form $\mathcal{N}(\theta_a;\mu_{ta},\Sigma_{ta})$. Actions are selected at each time $t$ by sampling $\hat{\theta}_a$ from $\mathcal{N}(\theta_a;\mu_{ta},\Sigma_{ta})$ and choosing the action with the largest value $\hat{\theta}_a^\top \mathit{v}_t$. The prior distribution for $\theta_a$ is of the form $\mathcal{N}(\theta_a;\mu_{0a},\Sigma_{0a})$. The distribution over $\theta_a$ for the selected action is updated at time $t$ based on the observed reward $r_t$ and $\mathit{v}_t$ using Bayesian inference. We provide the update equations for the mean and covariance matrix below. 
\begin{align}
    \Sigma_{(t+1)a} &= \sigma_{Ya}^2  ~ \big( \mathit{v}_t^\top   \mathit{v}_t + \sigma_{Ya}^2 ~ \Sigma^{-1}_{ta}  \big)^{-1}\\
    \mu_{(t+1)a}   &= \Sigma_{(t+1)a}  ~ \big((\sigma_{Ya}^2)^{-1} ~ r_t ~ \mathit{v}_t +  \Sigma^{-1}_{ta} ~ \mu_{ta}  \big)
\end{align}\label{equations:TS posterior}

\noindent\textbf{Related work.} Recent works use LLMs in RL, where the RL agent selects actions based on natural language inputs, and apply to games \citep{Du2023}. Note that in our work we leverage LLMs as foundational models, and focus on online decision-making for episode-limited RL settings, thus while we combine LLMs and RL, our work differs from the recent research on RL from human feedback (RLHF) and from AI feedback (RLAIF), which typically require some form of reward modeling, and a large number of episodes to perform well. Other works have also explored using natural language inputs, but apply to recommender systems for items such as movies, or instructability of social media recommendation algorithms \citep{Lyu2024, Feng2024, Mysore2023, Sanner2023}. However, these approaches also require a large number of iterations to work well. In contrast, we use TS which is a Bayesian approach that can perform well in a lower number of iterations than typical deep RL methods.

\vspace{1em}

\section{Methods}
\label{sec: LLM RL Methods}

We start with an overview of our new method in Figure \ref{fig: Overview}, and summarize below how to frame an adaptive intervention as an RL system using LLM. Then we provide the details.

\begin{enumerate}

    \item Run a typical RL loop to select a candidate action $a$.

    \item Extract LLM decision: given the user preference and previous data, we construct the prompt, send it to the LLM, then extract the LLM decision from the LLM response.

    \item Update RL policy based on LLM decision: we choose to either select $\tilde{a} = 0$ (no message) or send $\tilde{a} = a$ (set the hybrid action $\tilde{a}$ to the RL candidate action $a$). 
\end{enumerate}

Note: if the RL agent selects action $a > 0$ (indicating a candidate message to be sent) and a user preference is generated, then the LLM is prompted to decide if this message should actually be sent or not. If the RL agent selects action $a=0$ (indicating no message) or if there is no user preference that was generated, then there is no need to call the LLM, so the typical RL loop continues as usual.

To evaluate our method, we run our new simulation environment that generates text-based user preferences. 

Importantly, the RL loop and the user preference generation process are two separate loops. The text-based user preference is generated by the simulation environment, as described in Section \ref{sec: create StepCountJITAI for LLM}. The user preference is inserted into the LLM prompt, when a call to the LLM is made.

\subsection{StepCountJITAI for LLM}
\label{sec: create StepCountJITAI for LLM}

We extend the base simulator for adaptive intervention, introduced in \cite{karine2024}, to create StepCountJITAI for LLM, a novel simulation environment that works with LLMs. We describe the base simulator as well as the behavioral dynamics in Section \ref{sec: background StepCountJITAI}.

Our simulation environment, StepCountJITAI for LLM, can generate text-based user preferences, and incorporates constraints that affect the behavioral dynamics. To illustrate how it works, we focus on the user preference ``cannot walk''. Other user preferences can be implemented in the same way as described in our work.

\vspace{.5em}

Below, we start with a summary of how to create StepCountJITAI for LLM, then provide details.

\begin{enumerate}
    \item Augment the simulation environment states with a new binary auxiliary variable $W$, with values: $0$ ``cannot walk'' or $1$ ``can walk''. Note that this variable $W$  is hidden for the RL agent (i.e., not observed by the RL agent).

    \item Implement the dynamics for $W$ using a Markov chain.

    \item Generate a text-based user preference when $w=0$, randomly chosen from a list of pre-defined user preferences.
\end{enumerate}

\vspace{.5em}

\begin{figure}[ht]
    \begin{center}            
        \begin {tikzpicture}[scale=0.55, ->, line width=0.5 pt, node distance=1.3cm]
            \node[circle,draw,thick,fill=gray,text=white](zero) {0};
            \node[circle,draw,thick,fill=gray,text=white](one) [right of=zero] {1};
            \path(zero) edge [loop left,red] node {$1-p_{w_{01}}$} (zero);
            \path(zero) edge [bend left,red] node[above] {$p_{w_{01}}$} (one);
            \path(one)  edge [bend left,blue] node[below] {$1-p_{w_{11}}$} (zero);
            \path(one)  edge [loop right,blue] node {$p_{w_{11}}$} (one);
        \end{tikzpicture}
    \end{center}
    \vspace{-1em}
    \caption{Markov chain sketch.}
    \label{fig: Markov chain sketch}
    \vspace{.3em}
\end{figure}
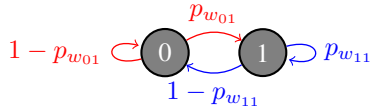

\noindent \textbf{Creating  auxiliary variable $W$ (cannot walk / can walk)}. We first augment the simulation environment with a new binary state variable $W$ with value: $0$ ``cannot walk'' or $1$ ``can walk''. The variable $W$ is not observed by the RL agent. It reflects a hidden state of the user, and is used to generate the user preference. 
We implement a Markov chain to simulate $w_t$, the values of $W$ at time $t$. The Markov chain and transition function for $W$ are shown in Figure \ref{fig: Markov chain sketch} and Table \ref{tab: Transition Function}.

\begin{table}[H]
\centering
    \begin{tabular}{c|c||c}
        $w_t$ & $w_{t+1}$ & $P(w_{t+1}|w_t)$ \\
        \hline
        0 & 0 & $1-p_{w_{01}}$ \\
        0 & 1 & $p_{w_{01}}$ \\
        1 & 0 & $1-p_{w_{11}}$ \\
        1 & 1 & $p_{w_{11}}$ \\
    \end{tabular}
    \caption{Transition Function.}
    \label{tab: Transition Function}
\end{table}

\noindent We define the new parameters: $p_{w_{01}}$ the probability of transitioning from $w_t=0$ to $w_{t+1} = 1$, and $p_{w_{11}}$ the probability of remaining in the ``can walk'' state.
\begin{align}
    p_{w_{01}} &= P(w_{t+1}=1|w_t=0)\\
    p_{w_{11}} &= P(w_{t+1}=1|w_t=1)
\end{align}

\noindent 

\vspace{.5em}

Setting $p_{w_{11}}$ to a lower (or higher) value allows for a lower (or higher) probability of remaining in the ``can walk'' state. Similarly, setting $p_{w_{01}}$ to a lower (or higher) value allows for a lower (or higher) probability of transitioning from $w_t=0$ to $w_{t+1} =1$. 

We note that the parameters $p_{w_{01}}$ and $p_{w_{11}}$ can be used to simulate the user state ``cannot walk'' over a variety of ranges, from shorter to longer time intervals, and thus enabling a variety of scenarios for our experiments.

\vspace{1em}

\noindent \textbf{Generating a text-based user preference ``cannot walk''.} Following the Markov chain and transition function in Figure \ref{fig: Markov chain sketch} and Table \ref{tab: Transition Function}, the variable $w_t$ can take values $0$ ``cannot walk'' or $1$ ``can walk''.

When transitioning from ``cannot walk'' to ``can walk'', the user preference is set to none, and the behavioral dynamics are not impacted by any constraints. The behavioral dynamics are the same as in the base simulator in \cite{karine2024}.

When transitioning from ``can walk'' to ``cannot walk'', a text-based user preference is randomly chosen from a list of pre-defined reasons for ``cannot walk''. This list was previously created by asking ChatGPT to give reasons why a user cannot walk. 

\vspace{.5em}

\noindent We show 20 reasons for ``cannot walk'':

\begin{small}
\noindent \texttt{\textcolor{blue}{I am tired, I do not want to walk, I got an injury, my leg is sore, I have a headache, the weather is bad, I have a cold, I feel unwell, I have a prior commitment, I have a blister, I’m feeling dizzy, I twisted my ankle, I am recovering from surgery, I need to rest, I have joint pain, I’m dealing with anxiety, I have a family obligation, I forgot my shoes, I don’t have anyone to walk with, I have to finish my work first}}.
\end{small}

\vspace{1em}

\noindent \textbf{Incorporating ``cannot walk'' constraint that impacts behavioral dynamics.}

When transitioning from ``can walk'' to ``cannot walk'', in addition to generating a user preference above, the additional ``cannot walk'' constraint affects the behavioral dynamics as below.

\vspace{1em}

\noindent\fbox{\begin{minipage}{.97\columnwidth}
\textbf{``cannot walk'' constraint:}
\begin{itemize}[nosep]
    \item disengagement risk $d_t$ is increased by $\eta_d \times d_t$

    \item habituation level $h_t$ is increased by $\eta_h \times h_t$

    \item reward is set to $0$.
\end{itemize}
\end{minipage}}

\vspace{1em}

\noindent The reward is the walking step count. It is set to $0$ since the user cannot walk. We introduce the constraint parameters $\eta_d$ and $\eta_h$, with values $\in [0,1]$. The base simulator variables and parameters are summarized in Appendix \ref{Appendix: JITAI simulation environment specifications}.

\vspace{1em}

\subsection{Construct LLM prompt}

We construct the LLM prompt by including a description of the mobile health app, the behavioral dynamics, the user current state and previous data, the user preference, and a question asking the LLM to make a decision ``send'' or ``not send'' a message to the user. 

Below, we provide an example of an LLM prompt.

\newpage

\noindent \textbf{Example of LLM prompt.} The LLM prompt contains the following blocks of text (a chain of instructions).

\vspace{.5em}

\begin{small}
\noindent \texttt{\textcolor{blue}{
A mobile health app can send a message to the user to encourage the user to walk.
\\
...\\
Sending a message causes the habituation level to increase.\\
Not sending a message causes the habituation level to decrease.\\
An incorrectly tailored message causes the disengagement risk to increase.\\
A correctly tailored message causes the disengagement risk to decrease.\\
...\\
\textcolor{purple}{The user current state and previous data...}\\
...\\
The user preference is \textcolor{purple}{"I twisted my ankle"}.\\
...\\
Should the mobile health app send a message to the user?
}}
\end{small}

\vspace{1em}

We detail the \begin{small}\texttt{\textcolor{purple}{text in purple}}\end{small}. The block of text contains the current user state and several previous rows of data (e.g., $h_t, d_t, r_t, c_{t-1}, h_{t-1}$,...). We experimented with various windows of previous data, and chose to insert the $4$ previous rows of data, since inserting more rows did not significantly impact the results. The block of text for the user preference (e.g., ``I twisted my ankle'') is chosen randomly from the list provided in Section \ref{sec: create StepCountJITAI for LLM}.

\vspace{1em}

\subsection{Select hybrid action based on RL candidate action and LLM decision}

Based on the given prompt, the LLM responds with a decision. The LLM decision is to either send a message to the user (i.e, $\tilde{a}=a$, where $a$ is the RL candidate action) or not send any message to the user (i.e., $\tilde{a}=0$). In other words, based on the prompt that includes the user preference and other data, the LLM can understand if a message should be sent or not, and thus can override the original candidate action selected by the RL agent. Note that the RL agent does not have knowledge of the user preference.

\vspace{.5em}

\noindent \textbf{Example of LLM response.} If the user preference is ``I twisted my ankle", then the LLM will recognize that a message should not be sent to the user. Thus, the LLM response is ``not send'', then the hybrid action is $\tilde{a} = 0$ (the original RL candidate action is discarded). Otherwise the LLM may decide to answer ``send'', in which case the hybrid action is $\tilde{a}=a$ ($\tilde{a}$ is set to the original RL candidate action). 


\section{Experiments}
\label{sec: LLM RL Experiments}

Below, we first describe how to validate the LLM response, then we describe our practical method for accelerating personalization in adaptive intervention. 

\subsection{Validating LLM response}

We perform experiments to check if the LLM response is correct. For each experiment, we generate prompts for ``cannot walk'', along with the constraint to impact the behavioral dynamics. Ideally, the LLM response to these prompts should be ``not send'' a message. We also ask the LLM to provide a reason for its decision. Below are some examples.

\vspace{1em}

\noindent \textbf{Example 1:}

\noindent The user preference is:
\begin{small}
\texttt{\textcolor{purple}{ my leg is sore.}}
\end{small}

\vspace{.5em}

\noindent The LLM correctly decides ``not send'' with the reason:
\begin{small}
\texttt{\textcolor{blue}{The user has a sore leg, indicating they cannot walk. The app should not send a message.}}
\end{small}

\vspace{.5em}

\noindent The LLM manages to provide a valid reason. This shows how an LLM can be a useful tool.

\vspace{1em}

\noindent \textbf{Example 2:}

\noindent The user preference is:
\begin{small}
\texttt{\textcolor{purple}{ I’m feeling dizzy.}}
\end{small}

\vspace{.5em}

\noindent The LLM correctly decides ``not send'' with the reason:
\begin{small}
\texttt{\textcolor{blue}{The user is feeling dizzy, which means they cannot walk, so the mobile health app should not send a message.}}
\end{small}

\vspace{.5em}

\noindent Again, the LLM manages to provide a valid reason.

\vspace{1em}

\noindent \textbf{Example 3:}

\noindent The user preference is
\begin{small}
\texttt{\textcolor{purple}{I am dealing with anxiety.}}
\end{small}

\vspace{.5em}

\noindent The LLM incorrectly decides ``send'', with the reason:
\begin{small}
\texttt{\textcolor{blue}{The user is expressing anxiety, and a tailored message could be helpful.}}
\end{small}

\vspace{.5em}

In a few cases, the LLM incorrectly decides to send a message, when the user cannot walk (i.e., the ``cannot walk'' constraint impacts the behavioral dynamics). Further investigation reveals that this occurs when the text-based user preference is ambiguous, thus does not clearly indicate if the user can or cannot walk. However, since these ambiguous text-based user preferences appear in less than $6\%$ of the time steps during our experiment, and since sending the hybrid action falls back to sending the RL candidate action, LLM+TS still outperforms the standard TS agent. 

Above, we have shown how to check if the LLM response is correct, thanks to our simulation environment, by tracking exactly where the LLM decision is incorrect. Future work would involve inserting additional insights into the LLM prompt to further improve the LLM response.

\subsection{Practical method for accelerating personalization in adaptive intervention.}

We conduct extensive experiments to compare our novel method LLM+TS to the standard TS. An experiment (a.k.a trial) corresponds to the behavioral study of one participant, where the maximum study length is $50$ days, with daily data.
We run our experiments for various combinations of the parameters $(p_{w_{11}}, p_{w_{00}})$, where $p_{w_{00}}$ = $1 - p_{w_{01}}$, to cover different scenarios. For example, the participant often sustains a light injury and thus often cannot walk for short periods, or the participant sometimes twists their ankle and thus sometimes cannot walk for longer periods. We repeat each experiment $10$ times with different seeds. 

For each experiment, we also run using various LLMs as foundational models, such as gemma and gemma 2, llama3 and llama 3.1, etc. \citep{gemma2024, Llama2024}. When using different LLM versions, we found that the choice of LLM version did not impact the results.

For each experiment setting, we compute the total reward as the sum of the rewards over a study. We use $\eta_{d} = \eta_{h} = 0.1$. The other experiment parameter values are provided in Appendices \ref{Appendix: JITAI simulation environment specifications} and \ref{Appendix: Thompson Sampling configurations}.


We present the results for two realistic scenarios: Scenario 1, where $p_{w_{11}} = 0.7$, and Scenario 2, where $p_{w_{11}} = 0.95$. In both scenarios, $p_{w_{00}}$ varies in the range $[0.1, ..., 0.5]$. Recall that $p_{w_{00}}$ is the probability of remaining in the ``cannot walk'' state, and $p_{w_{11}}$ is the probability of remaining in the ``can walk'' state.

\begin{figure}[t]
\begin{center}
\begin{subfigure}{\textwidth}
\includegraphics[width=0.18\textwidth]
{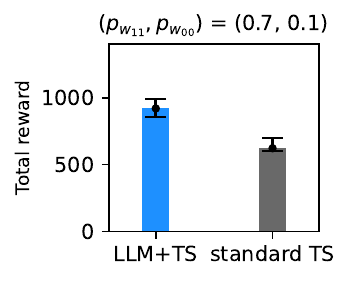}
\hfill
\includegraphics[width=0.18\textwidth]
{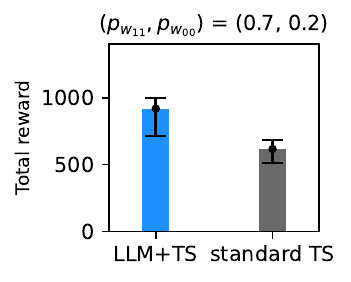}
\hfill
\includegraphics[width=0.18\textwidth]
{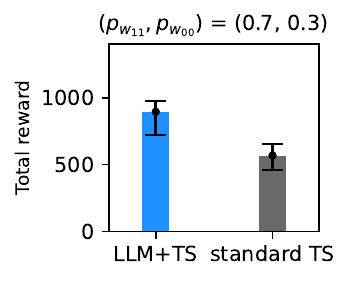}
\hfill
\includegraphics[width=0.18\textwidth]
{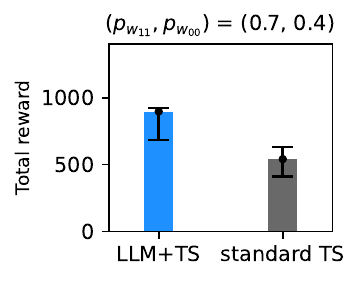}
\hfill
\includegraphics[width=0.18\textwidth]
{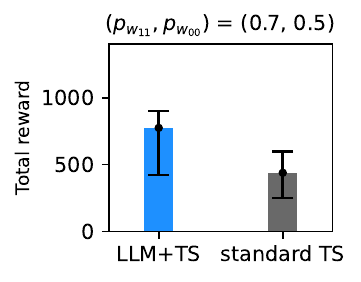}
\hfill
\vspace{-.5em}
\caption{Scenario 1: $p_{w_{11}} = 0.7$ (probability of staying in state ``can walk'') and various $p_{w_{00}}$ (probability of staying in  state ``cannot walk'').}
\vspace{1.5em}
\end{subfigure}
%
%
\begin{subfigure}{\textwidth}
\includegraphics[width=0.18\textwidth]
{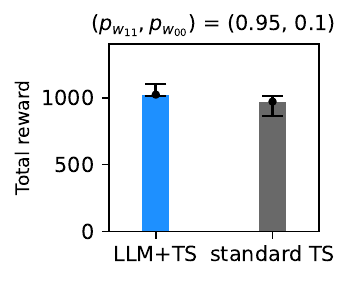}
\hfill
\includegraphics[width=0.18\textwidth]
{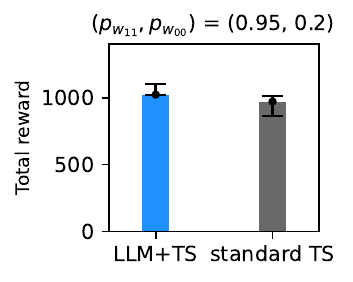}
\hfill
\includegraphics[width=0.18\textwidth]
{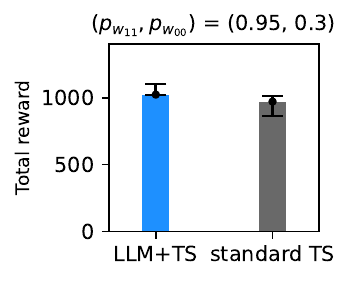}
\hfill
\includegraphics[width=0.18\textwidth]
{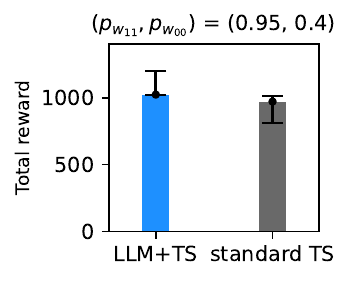}
\hfill
\includegraphics[width=0.18\textwidth]
{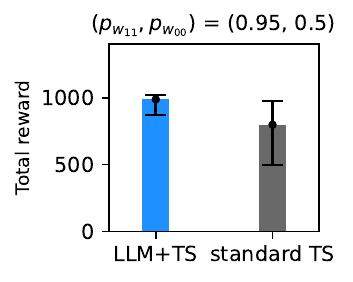}
\hfill
\vspace{-.5em}
\caption{Scenario 2: $p_{w_{11}} = 0.95$ (probability of staying in  state ``can walk'') and various $p_{w_{00}}$ (probability of staying in  state ``cannot walk'').}
\end{subfigure}
\end{center}
\caption{LLM+TS vs. standard TS example scenarios showing that LLM+TS  outperforms standard TS on most settings.}
\label{fig: MDP LLM+TS Scenarios}
\vspace{1em}
\end{figure}

In Figure \ref{fig: MDP LLM+TS Scenarios}, we plot the median total reward, along with the 25th and 75th percentiles, over all the trials. The plots show that LLM+TS, outperforms standard TS in most settings. 
For example, for $(p_{w_{11}}, p_{w_{00}}) = (0.7, 0.1)$, the median total reward for LLM+TS is $919.9$ (25th percentile is $852.6$ and 75th percentile is $990.4$), whereas the median total reward for standard TS is $622.5$ (25th percentile is $600.4$ and 75th percentile is $699.4$).


\begin{figure}[ht]
\begin{center}
%
%
\includegraphics[width=.6\columnwidth]{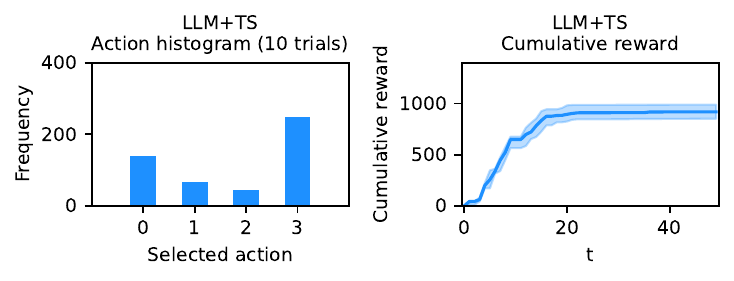}
\hfill
\includegraphics[width=.6\columnwidth]{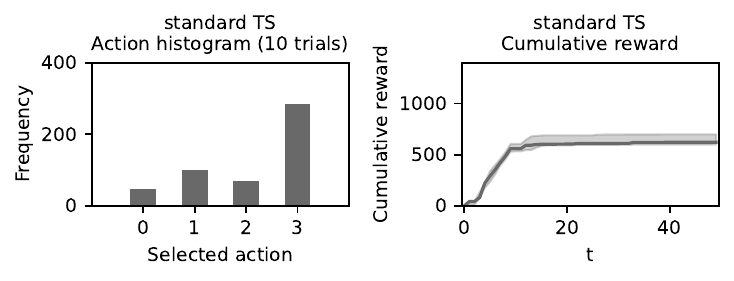}
\hfill
\end{center}
\caption{LLM+TS vs. standard TS. Example of histogram for all the selected actions, and plot of the cumulative rewards for $(p_{w_{11}}, p_{w_{00}}) = (0.7, 0.1)$. The histograms show that LLM+TS is able to capture a larger number of actions $0$, which indicates that the LLM has correctly decided to not send a message when the user cannot walk. The cumulative reward plots show that LLM+TS outperforms standard TS.}
\label{fig: hist actions and returns medians}
\vspace{1em}
\end{figure}

We compare the histograms of actions, taking into account all the selected actions across all the trials, for LLM+TS versus standard TS. We also compare the cumulative rewards. In Figure \ref{fig: hist actions and returns medians}, we show the histogram of all the selected actions, and the median cumulative reward, along with the 25th and 75th percentiles, over all the trials. We show that LLM+TS is able to capture a larger number of actions $0$, which indicates that the LLM has correctly decided to not send a message when the user cannot walk. This is further confirmed by the cumulative rewards for LLM+TS, which are higher than those for standard TS. Additional experiment results are provided in Appendix \ref{Appendix: additional action histograms}.

\vspace{1em}

\vspace{1em}


\section{Conclusion}
\label{sec: LLM RL Conclusion}

We introduce LLM+TS, a novel hybrid method that combines the LLM decision and the RL action selection, to update the RL policy, and thus to accelerate personalization in health adaptive intervention. We also introduce StepCountJITAI for LLM, a novel simulation environment that can be used to develop new RL algorithms for adaptive interventions using LLMs.
StepCountJITAI for LLM can generate text-based user preferences and incorporate constraints that affect the behavioral dynamics. Our results show that LLM+TS outperforms standard Thompson Sampling. Finally, we demonstrate how to frame a physical activity adaptive intervention as an RL system using LLM. Our method offers a promising solution framework for implementing a pipeline using LLM for personalized health adaptive interventions. 

\clearpage


\section*{Acknowledgements}
This work is supported by National Institutes of Health
National Cancer Institute, Office of Behavior and Social
Sciences, and National Institute of Biomedical Imaging
and Bioengineering through grants U01CA229445 and
1P41EB028242.


\bibliography{main}

\begin{thebibliography}{19}
\providecommand{\natexlab}[1]{#1}
\providecommand{\url}[1]{\texttt{#1}}
\expandafter\ifx\csname urlstyle\endcsname\relax
  \providecommand{\doi}[1]{doi: #1}\else
  \providecommand{\doi}{doi: \begingroup \urlstyle{rm}\Url}\fi

\bibitem[Agrawal and Goyal(2013)]{agrawal2013}
Shipra Agrawal and Navin Goyal.
\newblock Thompson sampling for contextual bandits with linear payoffs.
\newblock In \emph{Proceedings of the 30th International Conference on Machine Learning}, Proceedings of Machine Learning Research, pages 127--135, 2013.

\bibitem[Chu et~al.(2011)Chu, Li, Reyzin, and Schapire]{chu2011contextual}
Wei Chu, Lihong Li, Lev Reyzin, and Robert Schapire.
\newblock Contextual bandits with linear payoff functions.
\newblock In \emph{Proceedings of the Fourteenth International Conference on Artificial Intelligence and Statistics}, pages 208--214. JMLR Workshop and Conference Proceedings, 2011.

\bibitem[Coronato et~al.(2020)Coronato, Naeem, De~Pietro, and Paragliola]{coronato2020reinforcement}
Antonio Coronato, Muddasar Naeem, Giuseppe De~Pietro, and Giovanni Paragliola.
\newblock Reinforcement learning for intelligent healthcare applications: A survey.
\newblock \emph{Artificial Intelligence in Medicine}, 109:\penalty0 101964, 2020.

\bibitem[Du et~al.(2023)Du, Watkins, Wang, Colas, Darrell, Abbeel, Gupta, and Andreas]{Du2023}
Yuqing Du, Olivia Watkins, Zihan Wang, C\'{e}dric Colas, Trevor Darrell, Pieter Abbeel, Abhishek Gupta, and Jacob Andreas.
\newblock Guiding pretraining in reinforcement learning with large language models.
\newblock In \emph{Proceedings of the 40th International Conference on Machine Learning}, pages 8657--8677, 2023.

\bibitem[Feng et~al.(2024)Feng, Koo, Tan, Bruckman, and andAmy X.~Zhang]{Feng2024}
K.~J.~Kevin Feng, Xander Koo, Lawrence Tan, Amy Bruckman, and David W.~McDonald andAmy X.~Zhang.
\newblock Mapping the design space of teachable social media feed experiences.
\newblock In \emph{Proceedings of the CHI Conference on Human Factors in Computing Systems}, page 890–896, 2024.

\bibitem[{Gemma Team}(2024)]{gemma2024}
{Gemma Team}.
\newblock Gemma: Open models based on gemini research and technology.
\newblock \emph{arXiv:2403.08295}, 2024.

\bibitem[G{\"o}n{\"u}l et~al.(2021)G{\"o}n{\"u}l, Naml{\i}, Co{\c{s}}ar, and Toroslu]{gonul2021reinforcement}
Suat G{\"o}n{\"u}l, Tuncay Naml{\i}, Ahmet Co{\c{s}}ar, and {\.I}smail~Hakk{\i} Toroslu.
\newblock A reinforcement learning based algorithm for personalization of digital, just-in-time, adaptive interventions.
\newblock \emph{Artificial Intelligence in Medicine}, 115:\penalty0 102062, 2021.

\bibitem[Hardeman et~al.(2019)Hardeman, Houghton, Lane, Jones, and Naughton]{hardeman2019systematic}
Wendy Hardeman, Julie Houghton, Kathleen Lane, Andy Jones, and Felix Naughton.
\newblock A systematic review of just-in-time adaptive interventions (jitais) to promote physical activity.
\newblock \emph{International Journal of Behavioral Nutrition and Physical Activity}, 16\penalty0 (1):\penalty0 1--21, 2019.

\bibitem[Karine and Marlin(2024)]{karine2024}
Karine Karine and Benjamin~M. Marlin.
\newblock {StepCountJITAI}: simulation environment for {RL} with application to physical activity adaptive intervention.
\newblock In \emph{Workshop on Behavioral Machine Learning, Advances in Neural Information Processing Systems}, 2024.

\bibitem[Karine et~al.(2023)Karine, Klasnja, Murphy, and Marlin]{karine2023}
Karine Karine, Predrag Klasnja, Susan~A. Murphy, and Benjamin~M. Marlin.
\newblock Assessing the impact of context inference error and partial observability on {RL} methods for just-in-time adaptive interventions.
\newblock In \emph{Proceedings of the Thirty-Ninth Conference on Uncertainty in Artificial Intelligence}, volume 216, pages 1047--1057, 2023.

\bibitem[Liao et~al.(2020)Liao, Greenewald, Klasnja, and Murphy]{liao2020personalized}
Peng Liao, Kristjan Greenewald, Predrag Klasnja, and Susan Murphy.
\newblock Personalized heartsteps: A reinforcement learning algorithm for optimizing physical activity.
\newblock \emph{Proceedings of the ACM on Interactive, Mobile, Wearable and Ubiquitous Technologies}, 4\penalty0 (1):\penalty0 1--22, 2020.

\bibitem[{Llama Team}(2024)]{Llama2024}
{Llama Team}.
\newblock The llama 3 herd of models, 2024.

\bibitem[Lyu et~al.(2024)Lyu, Jiang, Zeng, Xia, Wang, Si~Zhang, Leung, Tang, and Luo]{Lyu2024}
Hanjia Lyu, Song Jiang, Hanqing Zeng, Yinglong Xia, Qifan Wang, Ren~Chen Si~Zhang, Chris Leung, Jiajie Tang, and Jiebo Luo.
\newblock Llm-rec: Personalized recommendation via prompting large language models.
\newblock In \emph{North American Association for Computational Linguistics}, 2024.

\bibitem[Mysore et~al.(2023)Mysore, Jasim, McCallum, and Zamani]{Mysore2023}
Sheshera Mysore, Mahmood Jasim, Andrew McCallum, and Hamed Zamani.
\newblock Editable user profiles for controllable text recommendations.
\newblock In \emph{Proceedings of the 46th International ACM SIGIR Conference on Research and Development in Information Retrieval}, pages 993--1003, 2023.

\bibitem[Nahum-Shani et~al.(2018)Nahum-Shani, Smith, Spring, Collins, Witkiewitz, Tewari, and Murphy]{nahum2018just}
Inbal Nahum-Shani, Shawna~N Smith, Bonnie~J Spring, Linda~M Collins, Katie Witkiewitz, Ambuj Tewari, and Susan~A Murphy.
\newblock Just-in-time adaptive interventions (jitais) in mobile health: key components and design principles for ongoing health behavior support.
\newblock \emph{Annals of Behavioral Medicine}, 52\penalty0 (6):\penalty0 446--462, 2018.

\bibitem[Russo et~al.(2018)Russo, Van~Roy, Kazerouni, Osband, and Wen]{russo2018}
Daniel~J. Russo, Benjamin Van~Roy, Abbas Kazerouni, Ian Osband, and Zheng Wen.
\newblock A tutorial on thompson sampling.
\newblock \emph{Found. Trends Mach. Learn.}, 11\penalty0 (1), 2018.

\bibitem[Sanner et~al.(2023)Sanner, Balog, Radlinski, Wedin, and Dixon]{Sanner2023}
Scott Sanner, Krisztian Balog, Filip Radlinski, Ben Wedin, and Lucas Dixon.
\newblock Large language models are competitive near cold-start recommenders for language- and item-based preferences.
\newblock In \emph{Proceedings of the 17th ACM Conference on Recommender}, page 890–896, 2023.

\bibitem[Thompson(1933)]{Thompson1933}
William~R. Thompson.
\newblock On the likelihood that one unknown probability exceeds another in view of the evidence of two samples.
\newblock In \emph{Biometrika}, volume~25, pages 285--294, 1933.

\bibitem[Yu et~al.(2021)Yu, Liu, Nemati, and Yin]{yu2021reinforcement}
Chao Yu, Jiming Liu, Shamim Nemati, and Guosheng Yin.
\newblock Reinforcement learning in healthcare: A survey.
\newblock \emph{ACM Computing Surveys (CSUR)}, 55\penalty0 (1):\penalty0 1--36, 2021.

\end{thebibliography}
\bibliographystyle{plainnat}

\clearpage


\appendix
\section{Appendix}

\vspace{1em}

\subsection{JITAI simulation environment specifications}
\label{Appendix: JITAI simulation environment specifications}

The base simulator introduced in \citep{karine2023} mimics a participant's behaviors in a mobile health study, where the interventions (actions) are the messages sent to the participant. We summarize the base simulator specifications in Tables \ref{tab:actions} and \ref{tab:env state}.

\begin{table}[h]
    \caption{Possible action values}
    \label{tab:actions}    
    \begin{center}
    \begin{tabular}{c@{\hskip 0.3in}l}
        \toprule
            \bf Action &\bf Description \\ 
        \midrule
            $a=0$    &  No message is sent to the participant. \\[1pt]
            $a=1$    &  A non-contextualized message is sent. \\[1pt]
            $a=2$    &  A message customized to context $0$ is sent. \\[1pt]
            $a=3$    &  A message customized to context $1$ is sent. \\[1pt]
        \bottomrule
    \end{tabular}
    \end{center}
\end{table}
\begin{table}[h]
    \caption{State variables}
    \label{tab:env state}
    \begin{center}
    \begin{tabular}{c@{\hskip 0.3in}l@{\hskip 
    0.3in}l}
    \toprule
        \bfseries Variable & \bfseries Description  & \bfseries Values\\
    \midrule
        $c_t$     & true context                  & $\{0,1\}$ \\
        $p_t$ & probability of context $1$    &  $[0,1]$\\
        $l_t$     & inferred context           & \{0,1\}\\
        $d_t$     & disengagement risk level      & $[0,1]$\\
        $h_t$     & habituation level             & $[0,1]$\\
        $s_t$     & step count               & $\mathbb{N}$\\[1pt] 
    \bottomrule
    \end{tabular}
    \end{center}
\end{table}

We describe the behavioral dynamics in Section \ref{Section Background: behavioral dynamics}. We use the same default parameter values as in the base simulator: context uncertainty $\sigma = 0.4$, behavioral parameters $\delta_h=0.1$, $\epsilon_h=0.05$, $\delta_d=0.1$, $\epsilon_d=0.4$, $m_s=0.1$, $\rho_1=50$, $\rho_2=200$. For our experiments, we set the disengagement threshold $D_{threshold} > 1.$. The maximum study length is 50 days with daily data. 

\vspace{1em}

\subsection{Thompson Sampling configurations}
\label{Appendix: Thompson Sampling configurations}
Using the same notations as in Section \ref{Section Background: Thompson Sampling}, we set the TS prior parameters $\mu_{0a}=0$ and $\Sigma_{0a}=100I$ for each action $a$, and the reward noise variance  $\sigma_{Ya}^2 = 25^2$ for each action $a$. 

\vspace{1em}

\subsection{Additional experiment results}
\label{Appendix: additional action histograms}

\begin{figure}[t]
\begin{center}
%
%
\begin{subfigure}{\textwidth}
\includegraphics[width=0.45\textwidth]{figures/experiments/ORIGINAL/fig2_hybrid_b_use_LLMTrue_pW11_pW00_0_7_0_1_llama3-70b-8192_trial9_episode0_prompt_obs_4_CHD.pdf}
\hfill
\includegraphics[width=0.45\textwidth]{figures/experiments/ORIGINAL/fig2_hybrid_b_use_LLMFalse_pW11_pW00_0_7_0_1_llama3-70b-8192_trial9_episode0_prompt_obs_4_CHD.pdf}
\hfill
\vspace{-.5em}
\caption{$(p_{w_{11}}, p_{w_{00}}) = (0.7, 0.1)$}
\vspace{1.5em}
\end{subfigure}
%
%
\begin{subfigure}{\textwidth}
\includegraphics[width=0.45\textwidth]{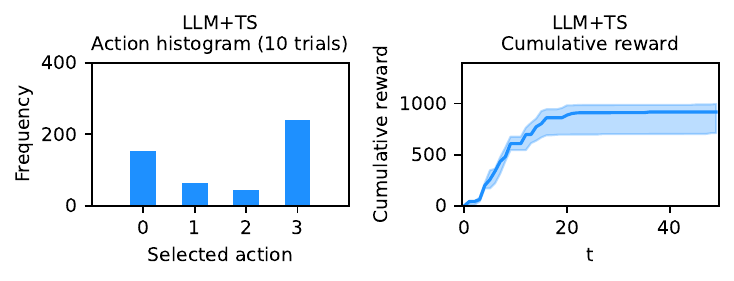}
\hfill
\includegraphics[width=0.45\textwidth]{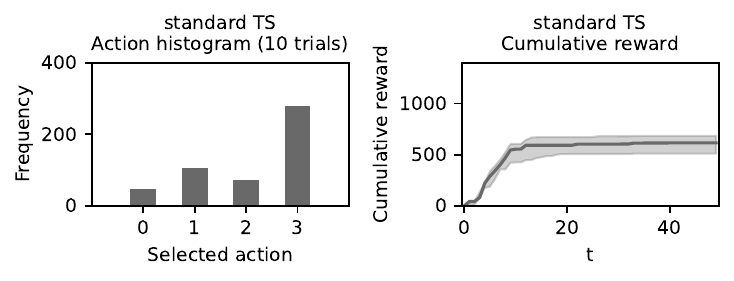}
\hfill
\vspace{-.5em}
\caption{$(p_{w_{11}}, p_{w_{00}}) = (0.7, 0.2)$}
\vspace{1.5em}
\end{subfigure}
%
%
\begin{subfigure}{\textwidth}
\includegraphics[width=0.45\textwidth]{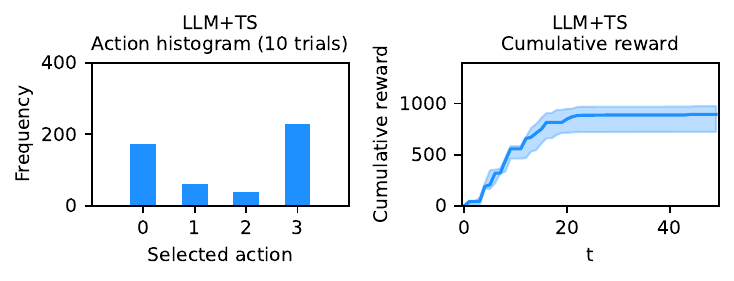}
\hfill
\includegraphics[width=0.45\textwidth]{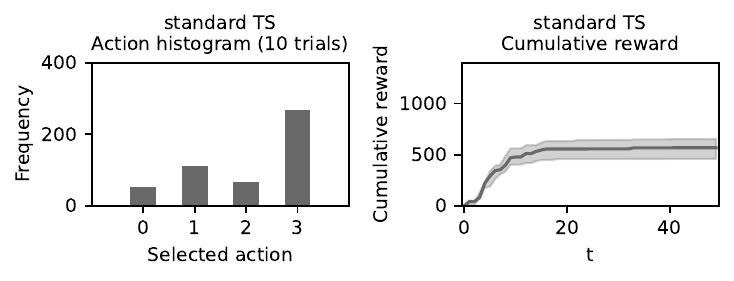}
\hfill
\vspace{-.5em}
\caption{$(p_{w_{11}}, p_{w_{00}}) = (0.7, 0.3)$}
\vspace{1.5em}
\end{subfigure}
%
%
\begin{subfigure}{\textwidth}
\includegraphics[width=0.45\textwidth]{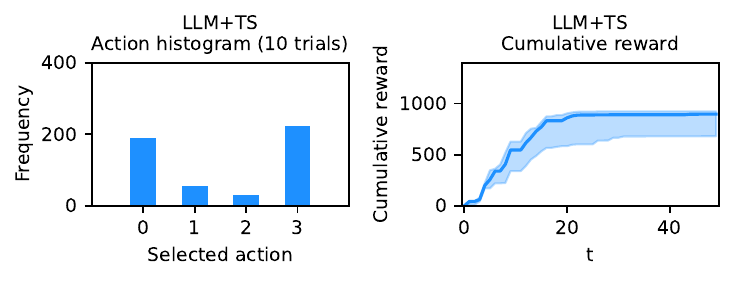}
\hfill
\includegraphics[width=0.45\textwidth]{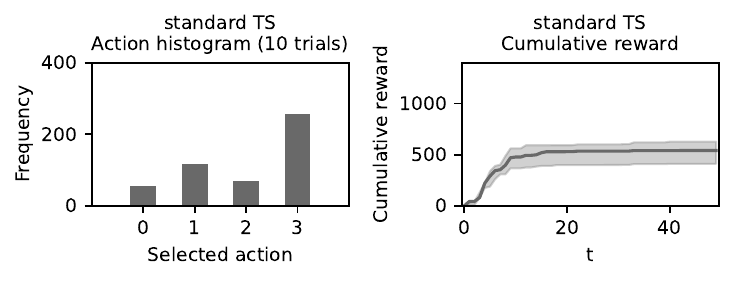}
\hfill
\vspace{-.5em}
\caption{$(p_{w_{11}}, p_{w_{00}}) = (0.7, 0.4)$}
\vspace{1.5em}
\end{subfigure}
%
%
\begin{subfigure}{\textwidth}
\includegraphics[width=0.45\textwidth]{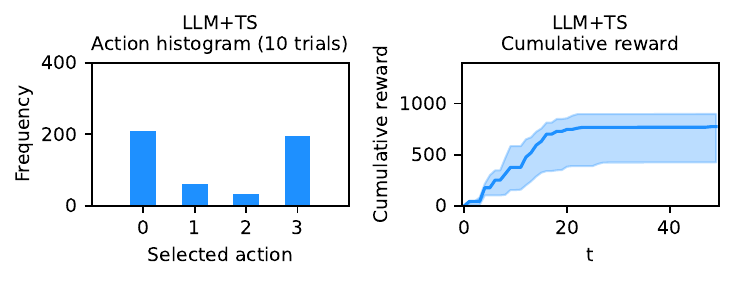}
\hfill
\includegraphics[width=0.45\textwidth]{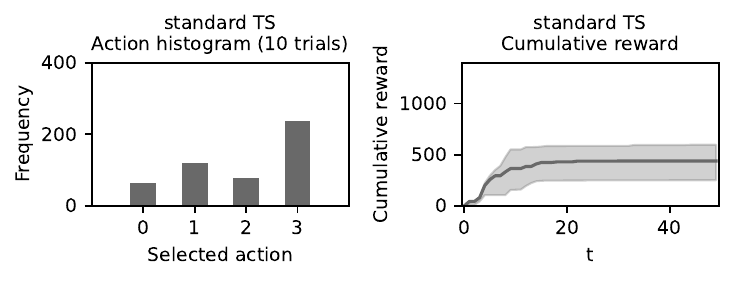}
\hfill
\vspace{-.5em}
\caption{$(p_{w_{11}}, p_{w_{00}}) = (0.7, 0.5)$}
\vspace{1.5em}
\end{subfigure}
%
%
\end{center}
\caption{LLM+TS vs. standard TS. Example of histogram for all the selected actions, and plot of the cumulative rewards for various $(p_{w_{11}}, p_{w_{00}})$ with fixed $p_{w_{11}}=0.7$ and varying $p_{w_{00}}$, when using LLM+TS (blue) and standard TS (gray). The histograms show that LLM+TS is able to capture a larger number of actions $0$, which indicates that the LLM has correctly decided to not send a message when the user cannot walk. The cumulative reward plots show that LLM+TS outperforms standard TS.}
\label{fig: hist actions and returns scenario 1}
\end{figure}

We run similar experiments as in Section \ref{sec: LLM RL Experiments} for various combinations of $(p_{w_{11}}, p_{w_{00}})$. We show the histogram of all the selected actions, and the median cumulative reward, along with the 25th and 75th percentiles, over all the trials, in Figure \ref{fig: hist actions and returns scenario 1}. The histograms show that LLM+TS is able to capture a larger number of actions 0, which indicates that the LLM has correctly decided to not send a message when the user cannot walk. The cumulative reward plots show that LLM+TS outperforms standard TS.

\end{document}